%% file: icat.tex
\documentclass[letterpaper, 10 pt, conference]{ieeeconf}  

\usepackage{algorithm}
\usepackage{algpseudocode}
\usepackage{amsmath,amssymb,breqn}
\usepackage{lipsum}
\usepackage[normalem]{ulem}
\useunder{\uline}{\ul}{}
\usepackage{makecell}\usepackage{color}
\usepackage{subfigure}
\usepackage{graphicx}
\usepackage[    colorlinks=true,
    linkcolor=blue,
    filecolor=blue,      
    urlcolor=blue,
    citecolor=blue,]{hyperref}
\usepackage{pifont} 
\usepackage{booktabs} 
\usepackage{adjustbox}  
\usepackage[table]{xcolor}
\definecolor{LightOrange}{RGB}{255,200,150} 
\definecolor{color1}{rgb}{0.8, 0.1, 0.1} 
\definecolor{color2}{rgb}{0.1, 0.8, 0.1} 
\definecolor{color3}{rgb}{0.1, 0.1, 0.8} 

\IEEEoverridecommandlockouts                              

\title{\LARGE \bf
ICAT: An Indoor Connected and Autonomous Testbed for \\ Vehicle Computing
}

 \author{
     Zhaofeng Tian\(^{1}\), Yuankai He\(^{1}\), Boyang Tian\(^{1}\),     Ren Zhong\(^{2}\), Erfan Foorginejad\(^{2}\), and Weisong Shi\(^{1}\)
     \thanks{} 
     \thanks{\(^{1}\)The CAR Lab, University of Delaware, Newark, USA
             {\tt\small \{zhaofeng, willhe, tby, weisong\}@udel.edu}} 
     \thanks{\(^{2}\)Department of Computer Science, Wayne State University,
             Detroit, USA
             {\tt\small \{ren, erfanf\}@wayne.edu}} 
}

\begin{document}

\maketitle
\thispagestyle{empty}
\pagestyle{empty}

\begin{abstract}
Indoor autonomous driving testbeds have emerged to complement expensive outdoor testbeds and virtual simulations, offering scalable and cost-effective solutions for research in navigation, traffic optimization, and swarm intelligence. However, they often lack the robust sensing and computing infrastructure for advanced research. Addressing these limitations, we introduce the Indoor Connected Autonomous Testbed (ICAT), a platform that not only tackles the unique challenges of indoor autonomous driving but also innovates vehicle computing and V2X communication. Moreover, ICAT leverages digital twins through CARLA and SUMO simulations, facilitating both centralized and decentralized autonomy deployments.

\end{abstract}

\begin{figure*}[htbp] 
\label{fig:panoptic}
\centering
\includegraphics[width=1\textwidth]{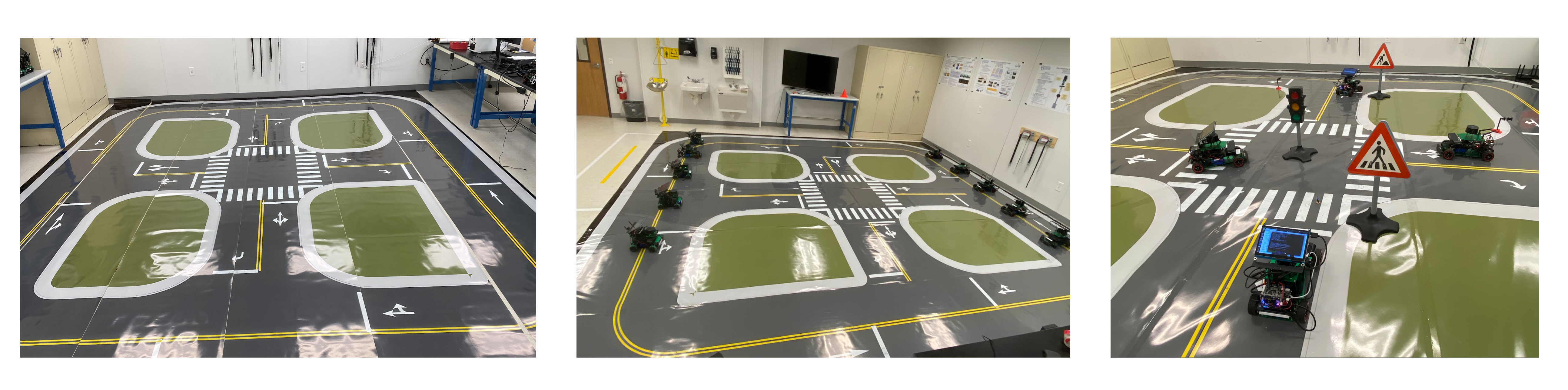} 
\caption{Panoptic view of ICAT platform in real world. ICAT is 6 by 5 meters in size, and 10 intelligent robots are used in ICAT now for autonomous driving studies. Traffic lights and signs are now integrated seamlessly with ICAT and enable interactions with traffic agents} 
\label{collision} 
\end{figure*}

\section{Introduction}

The emergence of autonomous driving technologies has necessitated the development of specialized testbeds for their validation and improvement. Prominent outdoor autonomous driving testbeds like Mcity ~\cite{mcity} at the University of Michigan, the edge computing testbed for autonomous driving at the University of Leeds ~\cite{edge_testbed}, Singapore's Smart Mobility Testbed ~\cite{NTU_smart_mobility}, and an autonomous shuttle testbed~\cite{shuttle_testbed} built upon Taltech smart mobility city in Estonia, have been instrumental in facilitating this progress. These environments replicate real-world traffic scenarios, offering invaluable data and insights. 

However, outdoor testbeds encounter several practical challenges that limit their efficacy and accessibility~\cite{indoor_cloud}. One of the most significant hurdles is the difficulty in creating realistic multi-agent traffic systems in a safe and controlled manner. Simulating dense, interactive traffic scenarios in outdoor settings poses substantial safety risks, making it challenging to test and refine autonomous systems under varied conditions. Additionally, the economic burden of establishing and maintaining such extensive outdoor infrastructure is substantial, often beyond the reach of smaller institutions and research entities.

Thanks to the more developed simulation technology, simulators like SUMO~\cite{sumo}, CARLA~\cite{carla}, and LGSVL~\cite{lgsvl} generate high-fidelity autonomous driving testing scenarios for the community while circumventing the financial costs of building an outdoor testbed. The gap between real-world autonomous driving and simulation, however, still exists because of the inability of the simulation to run hardware-in-loop vehicle computing and V2X analysis. Besides the pure simulations, to tackle the efficiency dilemma in both financial and time costs of a real-world testbed, several indoor autonomous driving testbeds have been developed ~\cite{Bassam, indoor_cloud, libai_testbed, udssc, icar_swarm_testbed}. These testbeds are ingeniously downscaled to fit limited indoor spaces, simulating real-world scenarios. Utilizing motion-capturing systems, they provide centralized localization information directly from the scenario, bypassing decentralized sensor data processing. This streamlined approach significantly benefits research areas like path-trajectory planning, traffic optimization, and swarm intelligence. However, these testbeds are somewhat limited in their sensing and computing capabilities due to the absence of robust sensors and computing power, which leads to inefficiency in simulating real-world decentralized commuting-intensive autonomous driving tasks.

We introduce ICAT, an Indoor Connected and Autonomous Testbed (ICAT) in response to these limitations. ICAT stands apart by not only focusing on the unique challenges of indoor autonomous driving but also pioneering in vehicle computing and V2X communication. With digital twins, specifically leveraging the capabilities of CARLA and SUMO simulations, ICAT enables both centralized and decentralized autonomy deployments in simulated and real-world testbeds. 

\textbf{The merits of ICAT can be summarized into two-manifolds. } \emph{(i) Connectivity and Vehicle Computing:} Compared to other indoor autonomous driving testbeds, ICAT is emphasized on its V2X capability, which supports inter-vehicle, vehicle-infrastructure, and vehicle-server communications. While the on-board computing devices could serve as the resources for handling data from vehicles, infrastructure, and other information systems. \emph{(ii) Digital Twin and Multi-agent Simulation:} The previous studies although built up a real-world multi-agent traffic or swarm system, faster developing process is prohibited without the full endorsement of a digital twin system. ICAT is not only integrated with CARLA, and SUMO interfaces but also a tailored pure Python multi-agent simulation environment, which releases the full potential in algorithm testing and iterating of autonomous driving tasks.

The rest of the paper is arranged as follows. The motivation and challenge of building ICAT are explained in Section~\ref{sec:motivation}, related works are introduced and compared in Section~\ref{sec:related}, and methods and technologies leveraged for designing ICAT are explained in Section~\ref{ICAT}. Two concrete case studies are conducted in Section~\ref{sec: experiment} for validating the ICAT system. The design and implementation experience is then discussed in Section~\ref{sec:discussion}, and conclusions are drawn in Section~\ref{sec:conclusion}.

\section{Motivation and Challenge}
\label{sec:motivation}

To better circumvent the unaffordable costs of building the outdoor autonomous driving testbed, and create more opportunities for the increasing interest in autonomy technologies, an indoor substitute would be an alternative to achieve the beginning stage of developing autonomous algorithms and modular structures. Moreover, thanks to the fast iteration of high-fidelity simulation platforms, developers could test their design in simulations conveniently.  Although some indoor autonomous driving testbeds have been built, we have observed some challenges in previous studies building such a system, which are explained below.

\emph{(i) Localization Accuracy:} An indoor testbed is highly downscaled compared to the real one, such that the localization accuracy in the downscaled road system plays a significant role in the downstream planning and control tasks. In some previous works, the localization is handled by a centralized motion-capturing vision system to extract the pose of each robot, such a method could limit the localization error to a smaller range of about 2cm. However, such a camera system for motion capture is very expensive, which brings more challenges for resource-limited entities.  Using a decentralized method with sensors like 2D lidars to localize the robot is more affordable but introduces more challenges in accuracy. 

\emph{(ii) Lack of On-board Computing Power:} In the current studies, most robot agent running in the testbed does not have a proper computing device for decentralized autonomous driving, whereby the planning module would solely rely on a center computer or a server. This limits the research on vision and machine learning, which is inefficient in supporting decentralized information extraction and planning for each intelligent agent.

\emph{(iii) Digital Twin with Simulation Capability:} The merit of a simulation platform for autonomous driving has been illustrated above. How to build a digital twin with access to such simulation software like CARLA and SUMO is vital for improving the quality and efficiency of the testbed. Unfortunately, the previous works do not support CARLA or SUMO interfaces.

To tackle these problems for a newer generation indoor autonomous driving testbed, we carefully choose different methods to solve them. The methods are explained in Section~\ref{ICAT} in detail.



\begin{table*} 
	\caption{Testbeds benchmark}
	\label{table1}
	\begin{adjustbox}{width=0.99\textwidth,center}
		\begin{tabular}{cccccccccc}
			\toprule
			\makecell[c]{\textbf{Research}} & \makecell[c]{\textbf{Centralized}} & \makecell[c]{\textbf{Decentralized}} & \makecell[c]{\textbf{Independent}\\ \textbf{Localization}}  & \makecell[c]{\textbf{V2X}\\ \textbf{Communication}}& \makecell[c]{\textbf{Vehicle}\\ \textbf{Computing}} & \textbf{Infrastructure}  & \makecell[c]{\textbf{Digital} \\ 
				\textbf{Twin}} & \textbf{Multi-agent} & \makecell[c]{\textbf{Carla /} \\ \textbf{SUMO}}  \\
			\midrule
			Vedder et al.~\cite{sweden_car} & \ding{51} & - & \ding{51} & - & - & - & - & - & - \\
			Tian et al.~\cite{tian} & - & \ding{51} & \ding{51} & - & \ding{51} & - & \ding{51} & - & - \\
			Ruch et al.~\cite{itsc_transporation_sim} & \ding{51} & - & - & - & - & - & - & \ding{51} & - \\
			Dosovitskiy et al.~\cite{carla} & - & \ding{51} & \ding{51} & - & - & \ding{51} & - & \ding{51} & \ding{51} \\
			Krajzewicz et al.~\cite{sumo} & \ding{51} & - & - & - & - & \ding{51} & - & \ding{51} & \ding{51} \\
			Pickem et al.~\cite{icar_swarm_testbed} & \ding{51} & - & - & - & - & - & - & \ding{51} & - \\
			Paull et al.~\cite{duckietown} & \ding{51} & - & - & - & - & \ding{51} & \ding{51} & \ding{51} & - \\
			Tran et al.~\cite{indoor_cloud} & \ding{51} & - & - & - & - & - & - & \ding{51} & - \\
			Li et al.~\cite{libai_testbed} & \ding{51} & - & - & - & - & - & \ding{51} & \ding{51} & - \\
			Scheffe et al.~\cite{Bassam} & \ding{51} & \ding{51} & - & - & \ding{51} & - & \ding{51} & \ding{51} & - \\
			Stager et al.~\cite{udssc} & \ding{51} & - & - & - & - & \ding{51} & \ding{51} & \ding{51} & - \\
			\textbf{ICAT} & \textbf{\ding{51}} & \textbf{\ding{51}} & \textbf{\ding{51}} & \textbf{\ding{51}} & \textbf{\ding{51}} & \textbf{\ding{51}} & \textbf{\ding{51}} & \textbf{\ding{51}} & \textbf{\ding{51}} \\
			\bottomrule
		\end{tabular}
	\end{adjustbox}
\end{table*}

\section{Related Work}
\label{sec:related}

In the domain of autonomous driving testbed, recent research has predominantly explored either simulation-based environments or real-world instances with limitations in scalability and decentralization. Studies like those in \cite{sweden_car} and \cite{tian} have focused on single-vehicle testing. The former developed a system using RTK localization and AUTOSAR design on a remote-controlled car, effective for controller design testing but lacking in broader traffic system impacts. The latter introduced an autonomous delivery robot with a Gazebo environment, supporting real-world simulation collaborations, yet not addressing multi-agent autonomous driving capabilities.

On the pure traffic simulation front,  AMoDeus, a testbed for autonomous transportation analysis is introduced in \cite{itsc_transporation_sim}, which is confined to simulations without real-world applicability or sensory integration. Popular simulation platforms CARLA and SUMO enable 3D and 2D traffic generation and partial sensing capabilities. Both simulators have rich interfaces open to the public, however, they still provide sim-to-real transferring with limited resources. 

Studies focusing on multi-robot systems, like \cite{icar_swarm_testbed} and \cite{duckietown}, have made strides in swarm formation and autonomous driving systems. The former uses QR tags for localization but omits autonomous driving systems and onboard computing. The latter, despite including traffic signs and a digital twin map, is constrained by its design for cost-effectiveness, limiting independent localization and computing.

Research in indoor autonomous driving testbeds, such as in \cite{indoor_cloud, libai_testbed, Bassam, udssc} utilized motion-capturing systems for localization. Robots are controlled via a cloud server but are not equipped with independent driving capabilities in \cite{indoor_cloud}. Study~\cite{libai_testbed}, while setting up a lane-free environment for multi-agent control algorithm testing, similarly did not enable decentralized methods. Scheffe et al.~\cite{Bassam} introduces a scaled, remotely accessible autonomous driving environment. It supports multi-agent MPC-based collision avoidance in a centralized manner using a motion-capturing system. Nonetheless, it does not address decentralized autonomous driving with multi-sensors. Lastly, Stager et al.\cite{udssc} describe a similarly scaled, remotely accessible autonomous driving environment, including simulated buildings. While it allows multi-agent centralized control with motion-capturing localization, the system is not designed for decentralized information processing and computing.

These works highlight a trend toward developing a scalable, decentralized autonomous driving testbed with onboard communication and computing capabilities. A brief comparison between our ICAT testbed and previous works is attached in Table~\ref{table1} where independent localization shows that the localization relies on the onboard sensors, not a centralized motion-capturing system. 
\begin{figure*}[htbp] 
\centering
\includegraphics[width=1\textwidth]{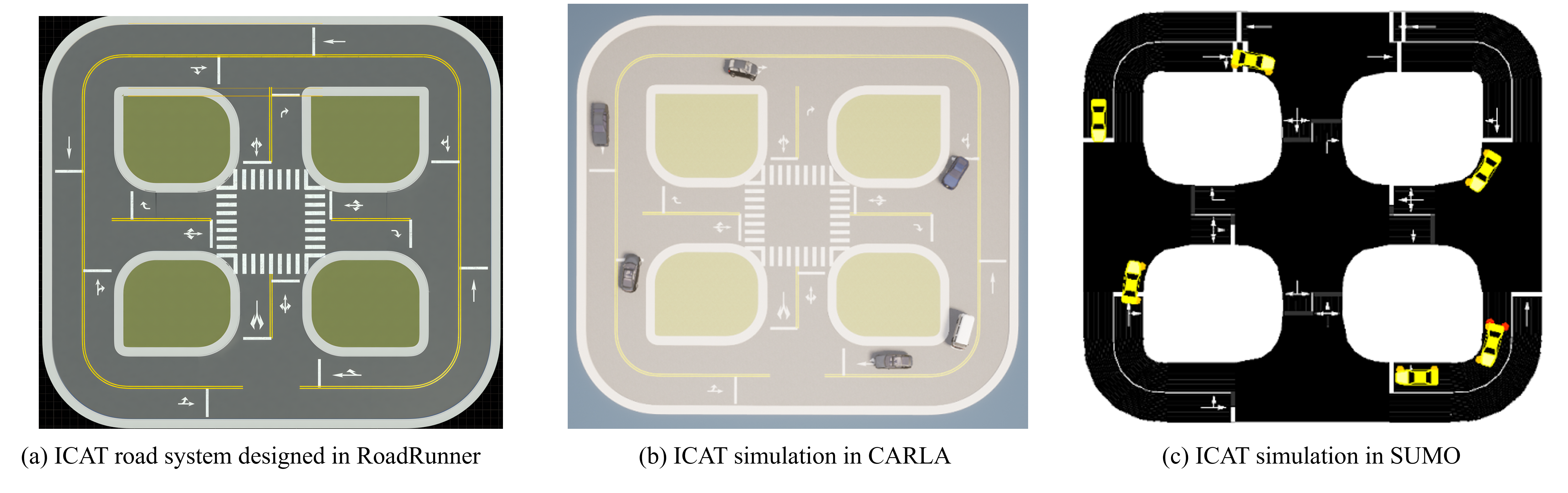} 
\caption{Digital twin system of ICAT. (a) ICAT's map is designed within RoadRunner~\cite{roadrunner}, which is a 3D interactive autonomous driving system builder. (b) The designed map then can be seamlessly imported into CARLA simulations. (c) SUMO simulations are also enabled with the ICAT road system.} 
\label{fig:digital} 
\end{figure*}

\section{ICAT Design} \label{ICAT}

In this section, different aspects of our proposed ICAT platform are introduced in detail. The methods and technologies that we have investigated for designing ICAT are illustrated as follows.

\input{DigitalTwin}

\input{Infrastructure}

\input{Localization}
\input{TrafficMangementSystem}

\input{WApproach}

\begin{figure}[htbp]
    \centering
    \includegraphics[width=0.8\linewidth]{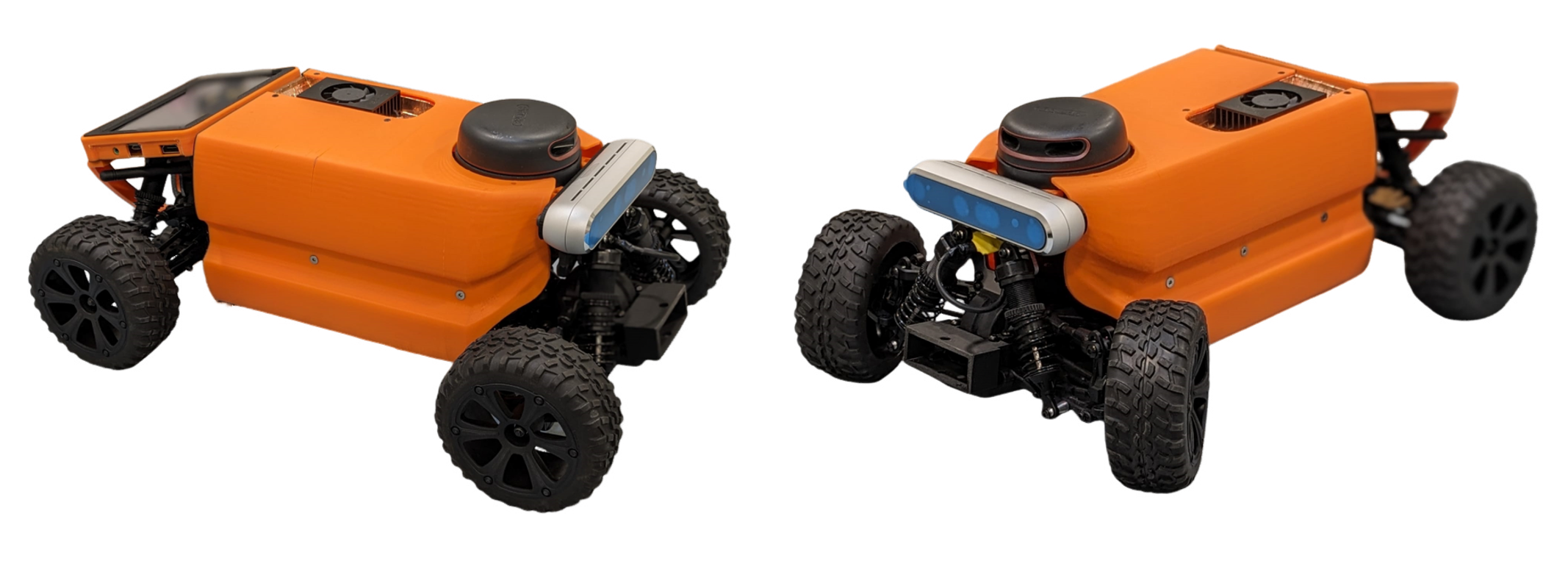}
    \caption{Our designed HydraT platform, is equipped with an advanced Lidar-camera sensing system and empowered with up-to-date navigation software. }
    \label{fig:HydraT}
\end{figure}

\input{VehicleComputing}

\begin{figure*}[htbp] 
\centering
\includegraphics[width=0.9\textwidth]{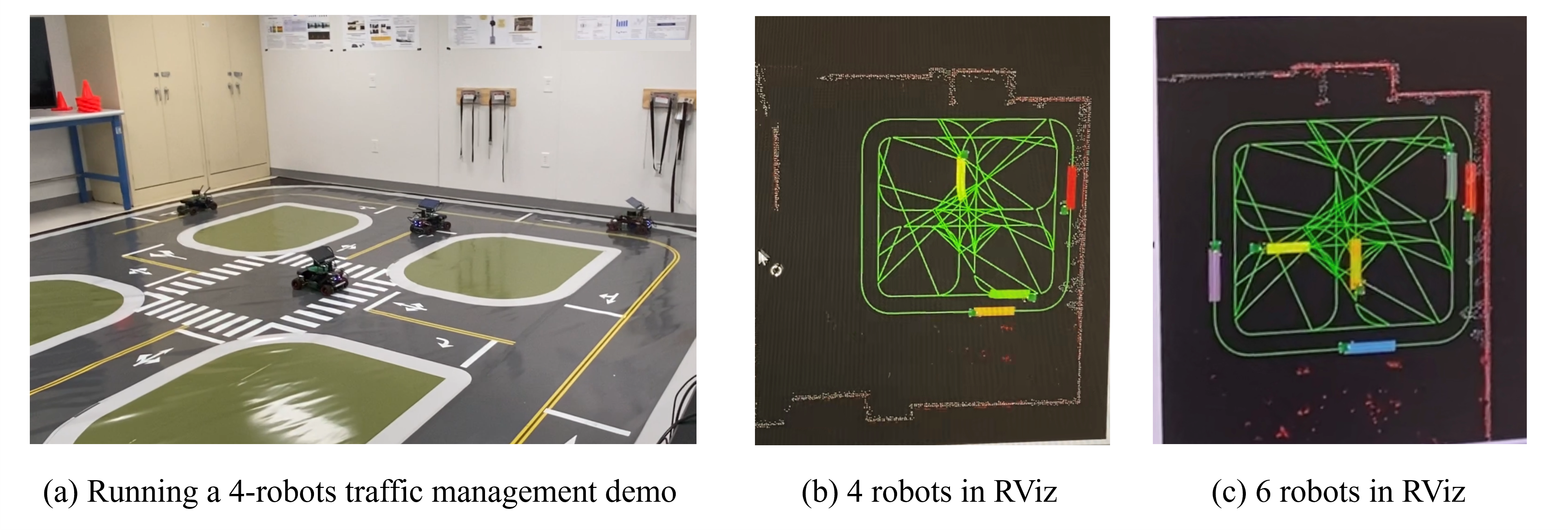} 
\caption{An illustration of traffic management real-world experiments. (a) 4 robots are running a traffic management demo. (b) Real-world demos can be also monitored in RViz, where the ICAT road is published as a Path message (the tilted lines are caused by the automatic connection of section start and end points in the Path message), the planned car trajectories from the central manager are rendered as different colors. (c) We have tested up to 6 robots in the ICAT real-world traffic system.} 
\label{fig: demo} 
\end{figure*}

\input{HydraT}

\section{Case Studies}
\label{sec: experiment}
The ICAT platform's efficacy is evaluated by two case studies: the operational integrity of the traffic management system and the execution of federated machine learning (ML) algorithms. These case studies are designed to validate the ICAT platform's functionality and showcase its versatility in handling complex computational tasks while managing a fleet of autonomous robots.

\subsection{Traffic Management System}
 As we explained in Section~\ref{sec: tms}, the whole management system is first built in a pure Python environment for validation. As shown in Fig.~\ref{fig:traj}, running 10 robots simultaneously in ICAT allows each car to automatically plan a path with random start and stop nodes with an A* algorithm. Then the car plans trajectories that dynamically adjust space and speed to a leading car in an ACC logic. The planned trajectory is rendered as apparent frames in the front of cars. 

However, in real-world experiments, the communications between cars and the centralized server are implemented in ROS~\cite{ros} system, which is different in simulations. To bridge this transferring gap between simulations and ROS real-world tests, we first assign the different robot namespace from $robot1$ to $robotn$ based on the number of robots used, hereby all the robot topics will be separated under their unique namespace. Next, different functions are separated into different ROS nodes, for instance, the localization module and trajectory planning module will publish poses and trajectories from different nodes. 

Besides interface issues, real-world validations, localization, and trajectory following (control) problems need to be handled carefully. This is because, in the simulation, the position/pose information is captured from the defined vehicle dynamic models without any sensors and control noises, which is assumed as ground truth. Assuming perfect trajectory tracking for every robot enables the robot to update its state perfectly to follow the trajectory, which is impossible in the real world where each frame in the trajectory necessitates accurate following by a well-designed control module. 

To respectively conquer the localization and control problem, we leveraged the NDT method~\cite{ndt} for localization and observed an accuracy improvement from the original AMCL method~\cite{amcl}, which has been explained in Section~\ref{sec: localization}. For the control method, we have implemented a pure pursuit method~\cite{pure_pursuit} to track the planned trajectory, where the controller takes a look-forward distance as input to chase the look-forward point, and hence improve the robustness against localization noises by ignoring the very close errors. 

To better visualize the real-world experiments and support monitoring at the digital end shown in Fig.~\ref{fig: demo}, we developed an RViz-based monitoring system where each car's trajectory and ICAT road system will be published as topics to be shown on screen. Furthermore, a pose initialization tool is integrated into the RViz, such that an initial pose for localization can be easily achieved by pulling the mouse toward the corresponding pose in the real world.

\subsection{Federated Machine Learning}
The second aspect of our evaluation focused on the execution of federated ML tasks. Each robot, during its simulated commute, participated in a distributed ML training session. The task involved collaboratively learning a predictive model while keeping on each robot the training data which is provided by our industry partner.

The experimental evaluation of the ICAT platform confirmed its robustness and versatility. The proposed traffic management system proved efficient, safe, and scalable. Simultaneously, the platform's capability to perform federated ML tasks without compromising its primary functions was successfully demonstrated. These results underscore the potential of ICAT as a dual-purpose platform, catering to the needs of intelligent traffic management and vehicle computing research.

\section{Discussion and Future Work}
\label{sec:discussion}

This paper presents the innovative ICAT platform and details our extensive experimentation and setup processes. Through our testing and evaluation, we have gained valuable insights into the complexities of autonomous vehicle technology. Our experiences and findings can be summarized in the following key observations:

\emph{(i) Impact of NDT Localization Noise on Pose Initialization. }
One of the significant challenges encountered during our experiments was the influence of noise in NDT localization on pose initialization. We observed that even minor noise in NDT measurements could lead to inaccuracies in initial pose estimation. This, in addition, also affected the navigation accuracy of the autonomous vehicles on our platform. The findings underscore the need for more robust noise filtering techniques in NDT localization to enhance the system precision.

\emph{(ii) Control Challenges in Trajectory Tracking. } Another key observation was related to the control aspects, particularly in tracking the planned trajectory using a pure-pursuit controller. While the pure-pursuit algorithm is a widely accepted method for trajectory following in autonomous vehicles, we noticed that the controller still exhibited errors in trajectory tracking, particularly on a curvy path. 

\emph{(iii) Effect of Communication Lag on Response Speed. } Lastly, our experiments brought to light the significant impact of communication lag on the response speed of the cars. The communication delay can have critical implications in a system where multiple autonomous vehicles communicate and operate synchronously. We observed that even minimal lags could lead to noticeable delays in the vehicles' responses to control commands. This finding is particularly pertinent in scenarios requiring fast decision-making and rapid maneuvering. 

In our future works, the insights we have obtained will be leveraged to improve the performance of the ICAT platform. Our future research plans will include the following items. First, optimize the localization accuracy with better filtering tricks. Second, a localization error-considered model predictive control method will be implemented to minimize the trajectory tracking errors. A learning method will acquire an accurate spatial-temporal environment dynamic model based on the collected real-time data. Finally, we will equip ICAT with more powerful communication devices, increasing the bandwidth at the hardware end. Furthermore, task scheduling and lag-considered safety protection will be investigated to avoid the harmful impact of communication delays.

\section{Conclusion}
\label{sec:conclusion}

This paper presented the ICAT platform, an advanced indoor testbed developed for cutting-edge research in autonomous vehicle systems and advanced vehicle computing. Through our evaluations, ICAT has demonstrated its proficiency in managing simulated traffic systems and executing complex federated ML tasks, proving its value as a versatile tool for the intricate demands of modern intelligent transportation and computational research.

A key strength of the ICAT platform lies in its integration of connectivity, vehicle computing, digital twin technology, and multi-agent simulation. The platform's exceptional V2X (Vehicle-to-Everything) capabilities and advanced onboard computing devices allow ICAT to adeptly manage and process data from various sources, including inter-vehicle, vehicle-infrastructure, and vehicle-server communications.

Furthermore, ICAT's digital twin system, integrated with CARLA and SUMO interfaces and a tailored pure Python multi-agent simulation environment, sets it apart from previous studies. This integration allows for the simulation of real-world multi-agent traffic or swarm systems and significantly accelerates the development and iteration of algorithms for autonomous driving tasks.

As we look to the future, we will focus on expanding ICAT's capacity, enhancing its simulation fidelity, and integrating decentralized computing paradigms more deeply. These advancements are directed towards supporting the evolving field of autonomous vehicle research, ensuring that ICAT remains a pivotal and invaluable resource for researchers and practitioners in this rapidly advancing area.

\section*{Acknowledgment}

The work is partially supported by the US National Science Foundation (NSF) under Grant No. 2311087. Special thanks are extended to Toyota and Tier IV for their support. The authors also wish to express their gratitude to all individuals and organizations that contributed directly or indirectly to the success of this work.

\bibliographystyle{IEEEtran}
\bibliography{main}

\end{document}

%% file: DigitalTwin.tex
\subsection{Digital Twin}

Digital Twin technology is a rapidly evolving field with numerous motivations and applications across various industries. In the autonomous driving industry, road system planning, traffic management, and machine learning can be facilitated by digital twins where the detailed and accurate representation of physical assets, systems, or processes, allows for better analysis, simulation, and decision-making. This is particularly useful in complex systems where physical testing is impractical or costly.

The merits of leveraging a digital twin for our proposed ICAT platform can be summarized in three aspects. \emph{(i) Algorithm testing.} Faster algorithm iteration can be achieved by efficient simulations. \emph{(ii) Traffic management. } Managing a group of interacted traffic agents with a centralized autonomous driving algorithm in the real world hardly guarantees safety, the counterpart in a digital twin can release such safety burden. \emph{(iii) Remote testing. } With advanced visualization capabilities, the information of testing in the real world can be accurately reflected on a remote end.

To build an indoor testbed for autonomous driving, a digital twin could support multi-purpose research and improve the developing efficiency. In such a digital design, however, we need to consider several problems. i)~What road structure or features should be included? ii)~ What size is feasible to be implemented in the real world? iii)~What interfaces it should have to integrate plug-ins of current autonomous driving simulators? 

The proposed problems are tackled in sequence for our design. i)~To meet the requirements of simulating intersections, merging points, and loops, we think a designed road system should include at least one intersection with left-turn, right-turn, and go-straight options for each direction. Also, merging points between lanes would support merging behavior research, and loops would simulate the traffic running in the system repeatedly with an automatic path planning algorithm. ii)~We need to make sure the size of the real-world system can fit into the lab room. So finally the real-world system is sized at 6 by 5 meters, which is a 10-times downscale of a 60 by 50 digital system. Thereby, the finalized design is processed in RoadRunner~\cite{roadrunner}, a road system building tool for autonomous driving development, and the final result is shown in Fig.~\ref{fig:digital}(a). Correspondingly, we printed the digital map and deployed it in the lab, which is shown in Fig.~1. iii)~To incorporate the convenience provided by current autonomous driving simulators, we build simulations using the designed ICAT map in both CARLA and SUMO, such that the built-in functions and algorithms can be leveraged by ICAT in a plug-and-play way. The CARLA and SUMO simulation demonstrations are shown in ~\ref{fig:digital}(b)(c), where the traffic could be either controlled by a traffic management algorithm from CARLA or SUMO. 

To support user-defined simulation scenarios, we have also developed a pure Python simulation system for ICAT, which enables the development of traffic management, trajectory planning, and control algorithms. One of the use cases is demonstrated in Fig.~\ref{fig:traj}, where a traffic management system with a trajectory planning algorithm is developed, and this process will be further illustrated in Section~\ref{sec: tms}

%% file: Infrastructure.tex
\subsection{Infrastructure and V2X}

ICAT also contains a down-scaled infrastructure system. To ensure a realistic representation, multiple traffic signs and traffic lights are used as Fig.~1 shown. The traffic signs include speed limit, yield sign, construction zone, stop sign, and do not enter signs.

The infrastructure information is also shared following the J2735 C-V2X standards. To accurately mimic realistic V2X communication, ICAT leverages the in-door WiFi system for communication. WiFi systems are similar to outdoor C-V2X systems due to their broadcasting nature. An underlying ROS communications stack is fitted into the WiFi system. A network model is used to mimic C-V2X message latency and packet loss.

Map information such as lane and intersection location and allow maneuvers are broadcasted by the infrastructure nodes. We've also added a Raspberry Pi controller to each traffic light signal so that traffic light state and timing information can be broadcast to nearby test vehicles. Furthermore, the J2735 standards also allow emergency vehicles to change traffic light states. This can be replicated with the help of the WiFi network, ROS communications stack, and the Raspberry Pi.


%% file: Localization.tex
\subsection{Localization}
\label{sec: localization}

 Compared to using an expensive centralized motion capturing system for localization, based on the onboard 2D Lidar, an ICAT robot can conduct localization using some mature algorithm in an efficient way. Additionally, a decentralized localization method could endorse richer testing scenarios for autonomous driving offering more realistic sensing inputs.

Specifically, to enable safe navigation of ICAT, we need to establish a solid localization foundation, including two functions, flexible global re-localization and real-time accurate pose tracking to produce reliable pose information of vehicle status. Global re-localization refers to obtaining the rough pose in the map with only the most recent observation when the localization system of ICAT is just booted or the tracking function fails. After producing an initial pose, the pose tracking function continues to estimate the movement of sensors periodically to ultimately calculate the vehicle status such as speed and orientation relative to the map.

Currently, there are many SLAM systems supporting both localization functions. For example, the ORB SLAM family \cite{orbslam1, orbslam2} which is a group of camera-based SLAM systems can locate the 3D world position of ORB features from images to track camera movements. By recording historical features into a library of historical pose, the ORB SLAM can achieve global re-localization by searching most similar features. However, this solution is less effective in an indoor environment as shown in a common indoor room as shown in the figure. Due to the limited performance of the camera depth sensor, the system found it hard to recognize the same wall at different distances. As a result, ORB SLAM is easy to commit tracking failure. Compared to camera-based SLAM, LiDAR-based SLAM \cite{scancontextloam, fastlio2} is better at pose tracking since the sensor data includes more accurate and wider depth information, but is worse at global re-localization due to the challenge of pose description especially for 2D LiDAR equipped by ICAT whose data is sparse and lack of geometrical structure.


In our implementation of the localization framework, we use Normal Distribution Transform (NDT) SLAM \cite{ndt} which is designed for 3D LiDAR originally to build a tracking layer due to its reliable performance compared to traditional 2D LiDAR SLAM such as \cite{gmapping, hector}. Furthermore, the generated layer is a 2D point cloud map that can be directly re-used by all robots in an ICAT fleet. NDT is also responsible for obtaining tracking results after obtaining the initial pose from global re-localization which is based on ORB features and historical pose library in the format of Bags of Words \cite{bagofwords}. Inspired by ORB SLAM, global re-localization will first extract ORB features from the latest image to calculate a visual word for searching the nearest word in the historical pose library. The found image is then matched with current features to estimate a rough pose. NDT SLAM can further optimize this pose to promise later tracking quality. 

%% file: TrafficMangementSystem.tex
\subsection{Centralized Traffic Management System}
\label{sec: tms}
\begin{figure*}[htbp] 
\centering
\includegraphics[width=1\textwidth]{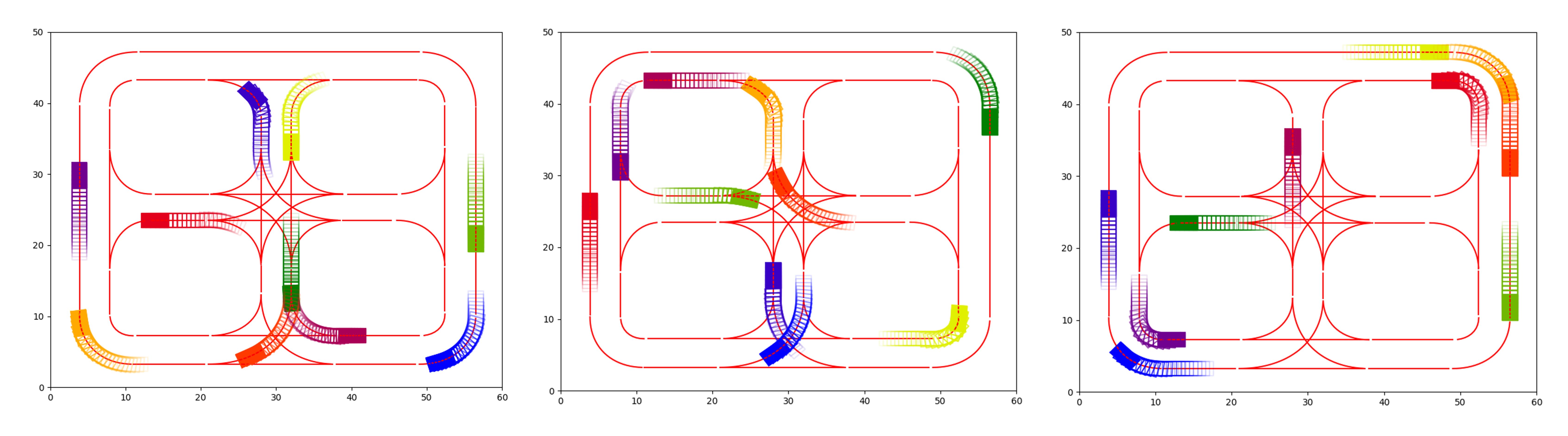} 
\caption{ICAT traffic management system testing in a Python simulation, where simulated 10 traffic agents could be controlled by a centralized traffic management algorithm. The planned trajectories for different robots are marked in different colors.} 
\label{fig:traj} 
\end{figure*}

CARLA and SUMO are designed primarily for autonomous vehicle research and development. Its traffic management system is created to simulate realistic traffic scenarios, providing a safe, controllable, and flexible environment for testing autonomous driving systems and algorithms. 

Both CARLA and SUMO primarily operate with a centralized architecture. In this system, the simulation environment, including traffic scenarios, vehicle behaviors, and environmental conditions, is controlled by a central server or application.
The control over individual vehicles, pedestrians, and other elements can be scripted or managed through the central API, allowing for precise control and manipulation of the simulation environment.
Specifically, in the context of autonomous vehicle testing, this centralized approach allows for a centralized planning framework and cooperative collision avoidance.

The merit of using a centralized traffic management system can be summarized as follows. i) Capable of simulating large networks of roads and junctions, complete with public transport, pedestrians, and varying traffic densities~\cite{andreas_nest}.
ii) Supports traffic light control, route planning, and vehicle-to-infrastructure communication.
iii) Can better support traffic flow analysis, such as measuring travel times, system throughout, and emissions~\cite{andreas_review}.

However, in the implementation of CARLA's traffic management system, we found a deadlock problem when the number of cars increased, and traffic efficiency degraded significantly. The reason is that in the intersections or merging points, the checking points are preset such that the first car hitting the checking point will go first leaving the other car totally stopped to yield. Such a First-In-First-Out(FIFO)~\cite{consensus} policy is not optimal, and further optimization is needed. 

In ICAT, we build such a traffic management system independently from CARLA and SUMO but refer to some ideas from them, i.e., the node-based traffic system graph. This graph will determine a full junction system of the roads, and support graph-based path search. Moreover, to tackle the problems of deadlocks and low efficiency, we leverage multi-agent trajectory planning to dynamically tune the speed for intersection/merging cars to pass at a higher efficiency.

\begin{algorithm}
\caption{Optimized Traffic Management System}
\label{alg: traffic manager}
\begin{algorithmic}[1]
\State Initialize traffic graph $G$, car states $X$, path buffer $P$, trajectory buffer $\text{Traj}$.
\State Initialize traffic signals, obstacles states $O$.
\For{each car $i$}
    \If{$P[i]$ is empty or invalid}
        \State $P[i] \gets \Call{A*}{G}$
    \EndIf
    \State $(s_i, d_i) \gets \Call{LocalizeToSD}{P[i]}$
    \State $\text{Traj}[i] \gets \Call{PolynomialPlanning}{s_i, d_i, X_i, O}$
    \State $\text{conflicts} \gets\Call{CollisionDetection}{\text{Traj}[i], \text{Traj}}$
    \For{each conflict in $\text{conflicts}$}
        \State $\text{Traj}[i] \gets \Call{ReplanTrajectory}{i, \text{conflict}}$
    \EndFor
\EndFor
\State Update global state based on new trajectories and environment changes for the next time step.
\end{algorithmic}
\end{algorithm}

\begin{figure}[htbp] 
\centering
\includegraphics[width=1\linewidth]{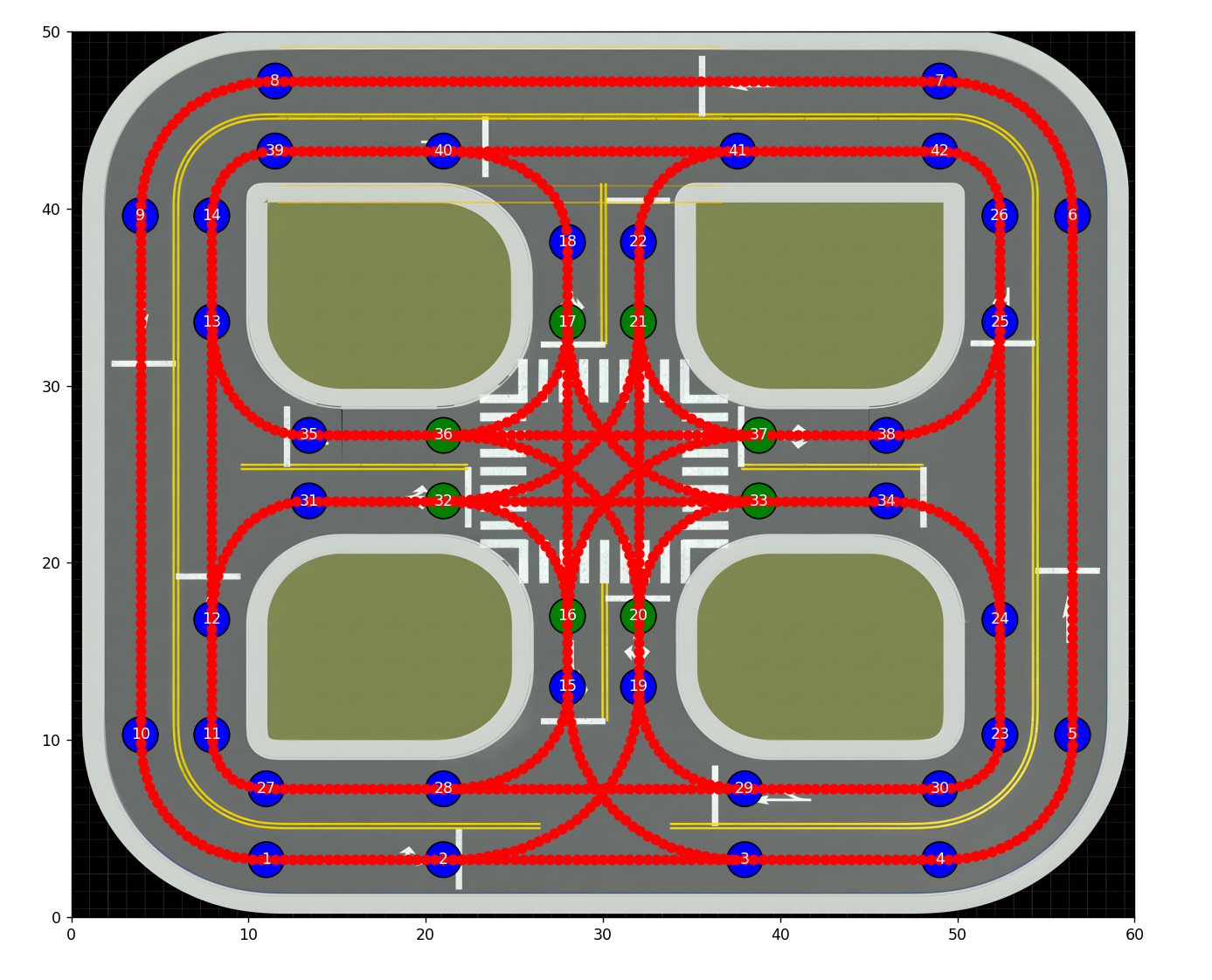} 
\caption{Graph topology for ICAT traffic system, consisting of nodes and edges that a pair of nodes is connected with either a straight line or a curve. Waypoints are discretized with a fixed distance as an attribute of the edge. } 
\label{fig:topo} 
\end{figure}

The process of operating such a traffic manager system is described in Algorithm~\ref{alg: traffic manager}. First, A traffic system graph is built up with nodes and edges, where each node is connected with others with a directed edge either shaped as a curve or straight line shown in Fig.~\ref{fig:topo}. Then A* search can be applied to each initialized robot to automatically generate paths and update goals (when approaching the previous goals) such that the simulation of a traffic system would not stop at reaching goals. 

After that, the coordinates of $(x,y)$ in Euclidian coordinates need to be transformed into $(s,d)$ Frenet coordinates~\cite{frenet}, such that the location of cars can be associated with roads in a convenient way. Based on the Frenet localization, the trajectory for each car is planned with polynomial interpolation between the current state and the future state. Other cars' states are also considered for trajectory planning, for example, the leading car's location and speed will be taken into account for doing an Adaptive Cruise Control (ACC) style trajectory planning. Other than ACC, the logic of trajectory planning for merging and diverging cases will be considered independently. A visualization of the planned trajectories is shown in Fig.~\ref{fig:traj}

Finally, the conflicts among planned trajectories will be addressed using a replanning process, where a non-linear optimization problem is usually set up to minimize the sum of the inter-car distance between conflicted cars for collision avoidance~\cite{distance}. The implementation of our designed traffic management system is illustrated in Section~\ref{sec: experiment}.

%% file: WApproach.tex
\subsection{Decentralized Autonomous Driving}

Decentralized Autonomous Driving differs from traffic management methods in that it doesn't rely on externally received command information, the traffic agents solely rely on their own sensors and computing devices to conduct perception and data processing. Thanks to heterogeneous on-board computing devices, the ICAT robots with ARM computing platforms leverage Scalable Open Architecture For the Embedded Edge (SOAFEE)~\cite{soafee} and Autoware~\cite{kato2018autoware} OpenAD Kit to support autonomous driving. SOAFEE and OpenAD Kit serve as a baseline for autonomous driving and enable researchers to quickly and easily modify the perception, planning, and control algorithms. Robot vehicles with other computing architectures leverage Autoware.AI, an autonomous driving stack based on ROS1, as the baseline.

Besides the mature autonomous driving pipelines like Autoware and SOAFEE, ICAT robots incorporate advanced vision reasoning technology for object detection and end-to-end navigation. Similar to Tesla's current solution, the current decentralized method uses depth cameras for environmental data perception~\cite{bevformer}. To ensure smooth driving along the road boundaries, we employ Color-based Lane Keeping~\cite{lane_detection}. The process begins with calibration, addressing distortions in the camera image caused by radial and tangential deformations. Calibration utilizes a known spatial relationship (calibration board) to deduce the camera's intrinsic parameters, correcting distortions in raw images.

Following calibration, we apply distortion correction to raw images and implement real-time HSV control for color tracking. Adjusting HSV thresholds filters out interfering colors, allowing for ideal recognition of squares in complex environments. Color transforms and gradients create a thresholded binary image.


Additionally, due to the limited variety and quantity of traffic sign shapes, we employ a lightweight YOLOv5 ~\cite{ultralytics2021yolov5} network for single-stage object detection to recognize these signs. We perform data augmentation by cropping standard traffic signs and introducing slight variations in brightness, angles, and horizontal flips, while simultaneously generating labeled bounding boxes. The lightweight YOLOv5 network helps reduce the computational load on the experimental vehicle, decrease space requirements, lower overall system energy consumption, and achieve real-time output speeds.

%% file: VehicleComputing.tex
\subsection{Vehicle Computing and Multi-user Management}
\label{sec:vehicle computing}

Besides the intelligent transportation and autonomous systems highlighted in the ICAT platform, the ICAT platform implements an innovative integration of vehicle computing and multi-user management. As the vehicle computing power is increasing rapidly, vehicle can serve as mobile computing platforms for various tasks.~\cite{vehicle_computing}. Adapting such an idea, our platform harnesses a fleet of 10 sophisticated robots with heterogeneous computing devices including Nvidia Nano, TX2, and NX. Each robot serves a dual purpose for computing: participating in a dynamic traffic system and acting as a resource for distributed computing tasks, such as federated Machine Learning (ML)~\cite{federated_ml}.

At the core of our platform lies the autonomous driving algorithms, which are rigorously tested within a simulated traffic system. For example, in decentralized autonomous driving, perception, and planning tasks will be processed onboard, where decentralized computational power plays an important role for autonomous agents to make decisions in real-time, adapting to the continuous flow of the traffic system.

Beyond traffic management, our robots with advanced computational capabilities, allowing them to engage in federated ML tasks during their ``commute". This innovative approach utilizes the otherwise idle processing power, turning each robot into a node within a distributed learning network. By doing so, we enhance the utility of our robotic fleet, contributing to a range of computational tasks for multi-user cases, from optimizing traffic algorithms to processing large-scale data analyses, all while ensuring the integrity and privacy of the data involved.

To address the complexities of multi-user management, our platform leverages Docker containers, providing a robust and isolated environment for each user. These containers encapsulate the necessary tools, libraries, and runtime for users to deploy and manage their autonomous driving and ML applications without interference. This separation is crucial for maintaining system stability and security, ensuring that the computational activities of one user do not adversely affect those of another.

%% file: HydraT.tex
\subsection{HydraT Platform and Auto-recharge}

Inspired by \cite{mushr, tian}, we developed a scaled autonomous car platform, Hydra-T, suitable for experiments in the ICAT environment shown as Fig.~\ref{fig:HydraT} While having the same Ackermann-steering system as a real car. To facilitate accurate sensing, the environment information can be extracted from both an accurate industry-level 2D-LiDAR, and an Intel RealSence depth camera. Combined with a 9-DOF IMU, the accuracy of localization and pose estimation can be further enhanced. Supporting intensive vehicle computation for multi-purposes, the robot is designed to adapt an Nvidia Orin machine that provides a powerful online machine inference capability.

To reduce maintenance requirements we design a custom charging dock for the robot, such that the robot can conduct automatic recharge before the battery is running out. This eliminates the need for researchers' intervention since the charging feature enables the robot to dock and recharge itself autonomously similar to vacuuming robots, thus achieving a higher time efficiency for a multi-agent system. Furthermore, the charging stations can also be a part of experiments simulating the real-world problems of electric vehicles with charging path planning and optimization for energy management.